\documentclass{article}

\usepackage{microtype}
\usepackage{graphicx}
\usepackage{subfigure}
\usepackage{booktabs} 
\usepackage{tablefootnote}

\usepackage{hyperref}

\usepackage[accepted]{icml2024}

\usepackage{amsmath}
\usepackage{amssymb}
\usepackage{mathtools}
\usepackage{amsthm}

\usepackage{todonotes}
\usepackage{graphicx}
\usepackage{subcaption}
\usepackage{tabularx}

\usepackage[capitalize,noabbrev]{cleveref}

\theoremstyle{plain}

\theoremstyle{definition}

\theoremstyle{remark}

\icmltitlerunning{Preprint}

\begin{document}

\twocolumn[
\icmltitle{A Comparative Study of Open Source Computer Vision Models \\ for Application on Small Data: The Case of CFRP Tape Laying}



\icmlsetsymbol{equal}{*}

\begin{icmlauthorlist}
\icmlauthor{Thomas Fraunholz}{cyber}
\icmlauthor{Dennis Rall}{wogra}
\icmlauthor{Tim K\"ohler}{wogra}
\icmlauthor{Alfons Schuster}{dlr}
\icmlauthor{Monika Mayer}{dlr}
\icmlauthor{Lars Larsen}{dlr}
\end{icmlauthorlist}

\icmlaffiliation{cyber}{Smart Cyber Security GmbH, Südportal 3, Norderstedt, Schleswig-Holstein, Germany}
\icmlaffiliation{wogra}{WOGRA AG, Hery-Park 3000, Gersthofen, Bavaria, Germany}
\icmlaffiliation{dlr}{German Aerospace Center (DLR), Am Technologiezentrum 4, Augsburg, Bavaria, Germany}

\icmlcorrespondingauthor{Thomas Fraunholz}{thomas.fraunholz@smart-cybersecurity.de}

\icmlkeywords{Machine Learning, Compute Vision}

\vskip 0.3in
]



\printAffiliationsAndNotice{}  

\begin{abstract}
In the realm of industrial manufacturing, Artificial Intelligence (AI) is playing a increasing role, from automating existing processes to aiding in the development of new materials and techniques. However, a significant challenge arises in smaller, experimental processes characterized by limited training data availability, questioning the possibility to train AI models in such small data contexts. In this work we explore the potential of Transfer Learning to address this challenge, specifically investigating the minimum amount of data required to develop a functional AI model. For this purpose, we consider the use case of quality control of Carbon Fiber Reinforced Polymer (CFRP) tape laying in aerospace manufacturing using optical sensors. We investigate the behavior of different open-source computer vision models with a continuous reduction of the training data. Our results show that the amount of data required to successfully train an AI model can be drastically reduced, and the use of smaller models does not necessarily lead to a loss of performance.
\end{abstract}

\section{Introduction}
\label{sec:introduction}
\subsection{From Big Data to Small Data: AI's New Frontier in Manufacturing}
\paragraph{} Artificial intelligence technology has become increasingly important in industrial manufacturing in recent years. They enable the automation of processes such as quality assurance, predictive maintenance and optimization of production processes. Since powerful AI models require a corresponding big amount of data, the focus at the beginning of this development was primarily on relatively few use cases that provided big data. Highly scaled processes, such as industrial series production, were particularly suitable for this purpose. Now that this area of application has reached a certain level of maturity, research has recently focused on the far more common applications where only a small amount of data is available. These applications can be found, for example, in the development of new materials and production processes, i.e. in the corresponding pre-series production. While the big data problems were about reducing the dataset by filtering to increase model quality, small data problems raise the question of whether there is even enough data to create an efficient AI model?

\subsection{Optimizing CFRP Tape Laying Processes}
\begin{figure}[ht]
	\centering
	\includegraphics[width=\columnwidth]{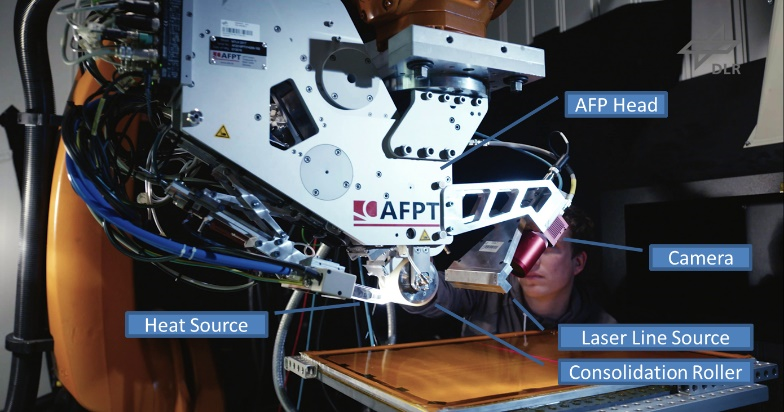}
	\caption{Automated Head for CFRP tape laying with optical sensor for quality control \cite{dlr137804}.}
	\label{fig:FPHead}
\end{figure}

\paragraph{} Let us illustrate this using the example of CFRP tape laying as shown in figure \ref{fig:FPHead}. CFRP tape laying is a versatile and efficient manufacturing method prominently utilized within the aerospace and automotive sectors. Its primary purpose is the construction of intricate and robust composite components, where lightweight, high-strength materials are essential. Its great versatility, allows manufacturers to produce parts with complex shapes and tailored mechanical properties. This new technology is on the edge of industrial maturity. This means that it operates in the area of pre-series production. Only small quantities will be produced under variable production processes, as these are subject to continuous optimization. This means that only little data is available for training AI models, such as for quality control. Nevertheless, automated and non-destructive testing is of great importance for these small batches in order to carry out further practical tests. Optical images are currently being taken continuously to detect component defects. The resulting images are further analyzed using complex algorithms. However, these have to be constantly adapted to the ever-changing process conditions during development, which represents a considerable effort. The challenge of this computer vision problem is to provide an automated quality control system that can adapt quickly to changing process conditions. A corresponding AI model would have to be able to adapt to the new process on the basis of a small number of examples classified by humans. Again, the question arises how many examples are necessary to create an AI model? 

\subsection{Related Work}
\paragraph{Transfer Learning in Computer Vision} Transfer learning is a common practice in the field of deep learning for computer vision. This technique involves creating high-quality models from large datasets and then fine-tuning them on smaller datasets \cite{Kolesnikov2019, Djolonga2020, dosovitskiy2020image}. These models perform significantly better than those trained solely on the smaller datasets. This approach leverages the knowledge gained from large datasets to enhance learning efficiency and accuracy on limited data. It has been widely adopted for computer vision tasks, including image classification, object detection, and, more recently, in complex scenarios such as autonomous driving and medical image analysis.

\paragraph{Deep Learning for CFRP} The deep learning method has already been investigated for automatic quality assurance in our CFRP use case \cite{Meister2021}. This exploration addressed the issue of limited data by utilizing Deep Convolutional Generative Adversarial Networks (DCGAN) and geometrical transformation techniques through synthetic image data augmentation. The resulting model is intended to serve as a benchmark for the quality of our fine-tuned models. However, we explicitly choose not to follow this path so that our findings can be applied to other use cases. Instead, we aim to determine under what circumstances and how much data are required to produce a comparable, if not superior, model using transfer learning.

\subsection{Our Approach}
\paragraph{Objectives and Contributions} We want to apply Transfer Learning for the application of optical quality control for the prediction of component defects in CFRP tape laying and show that transfer learning under the given small data leads to powerful models that outperform related solutions in this area. We restrict ourselves here to the use of open source pre-trained models. This should ensure that the present results can be adapted to similar applications and that the required computing capacities can also be applied to smaller development projects. In our comparative study, we address two key questions in particular. On the one hand, we want to compare different such pre-trained models as a basis for transfer learning. Thus, we want to find out what influence the model architecture and model size has in our use case. On the other hand, we address the question how a reduction of the data available for training affects the performance of the model. In doing so, we want to determine a lower limit for the amount of data required to create an AI model with the desired accuracy. 

\paragraph{Outline} This is done as follows: In section \ref{sec:tape-laying-process} we describe the use case from a technical point of view and in particular how the data for optical quality testing is generated and what properties it has. Next, in section \ref{sec:computational-methods}, we explain the methods used in transfer learning, i.e. from the choice of pre-trained models to the preprocessing of the data and the hyperparameter search used. The results obtained are presented and discussed in section \ref{sec:results}. Finally, we summarize the main aspects of our work in section \ref{sec:conclusion} and give an outlook on the resulting possibilities.

\section{CFRP Tape Laying}
\label{sec:tape-laying-process}
\paragraph{} Thermoplastic tape lay-up is a process where continuous strips of thermoplastic material, combined with reinforcing fibers such as carbon, are carefully placed onto a surface. This method is used to build layered composite structures. The tapes are often heated to their melting point to facilitate bonding between layers and the substrate.

\subsection{Horizontal Defects: Gaps and Overlaps}
\begin{figure}[ht]
	\centering
	\includegraphics[width=\columnwidth]{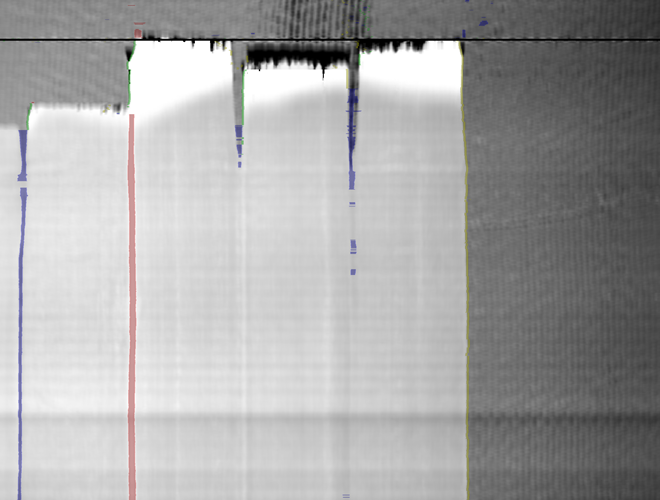}
	\caption{An exemplary high-resolution 16-bit TIFF image with annotation showing the measured height profiles at high contrast. The middle of the pictures shows the actual track consisting of three tapes arranged next to each other, displayed in light gray. To the left is the previous track, to the right the flat dark gray background. Each tape is 12,54 mm wide and may reach several meters in length, depending on the layup. The upper section of the image shows the beginning of the laid tape. On the left side in the previous track between the middle and right tape there is a gap (highlighted in blue), and at the border between the previous and the actual track there is an overlap (highlighted in red). Each of the 65,536 grayscale values of the original TIFF corresponds to a height differential of 1.77 micrometers, i.e., to represent the height difference of a tape, 79 grayscale values are required \cite{dlr137804}. }
	\label{fig:example-tiff}
\end{figure}

\paragraph{} During this process, a wide variety of possible defects can occur. In this work we concentrate us on two defects in horizontal direction which are gaps and overlaps, as shown in figure \ref{fig:example-tiff}. Gaps occur when there is insufficient material coverage between tape strips, while overlaps happen when tape strips excessively cover one another. Both defects may lead to potential structural weaknesses.

\subsection{Measurement Techniques in Automated Fibre Placement}
\paragraph{} Tapelaying and measurement are performed by a specialized robotic end-effector known as the automated fibre placement (AFP) head, as shown in figure \ref{fig:FPHead}. This head feeds multiple carbon fibre tapes to a heat source and then to a consolidation roller, where the material is firmly adhered to a jig of the desired final shape. Directly after the initial cool down, a tape placement sensor (TPS) acquires the height profile of the placed tapes by laser triangulation. Every profile, scanned at a distance of 400 microns, is added to a 16-bit gray scale TIFF image (figure \ref{fig:example-tiff}). The real world height profile may be obtained by scaling the TIFF-image with respect to the optical resolution in x-direction, the scan distance in y-direction and the sensor height calibration in z-direction. The previously described defects 'gap' and 'overlap' are then marked as mentioned in section \ref{sec:introduction}, the algorithm can be manually adjusted until the desired accuracy is reached. 


\subsection{Dataset Overview: Composition and Imbalance}
\paragraph{} In order to cover a sufficient defect size for the AI approach. Images of 152 pixels by 100 pixels were generated to cover each category. The pixels correspond to a physical image section of 6,2 mm by 40 mm. Due to a 0,4 mm resolution along the layup direction and 0,041 mm along the image width. Consequently, a dataset with 73,749 examples is formed, comprising 8,707 with a gap, 3,248 with an overlap, and 61,794 without defects, designated as nominal. The nominal category dominates with approximately 84\%, followed by gap with 12\% and overlap with 4\%. Thus, we are dealing here with an unbalanced dataset in which the nominal examples clearly dominate the two error categories. This corresponds to the real use case in industrial production, where a defect should be the exception and not the rule.

\begin{table*}[ht]
\centering
\caption{Overview of Selected Models with Corresponding Hyperparameters. }
\label{tab:models}
\begin{tabularx}{\textwidth}{lXrrrrr}
\toprule
&   &   & \textbf{Model} & \textbf{Batch} & \textbf{Learning} \\
\textbf{Short Name} & \textbf{Checkpoint}\footnotemark[1]  & \textbf{Reference} & \textbf{Size}\footnotemark[2] & \textbf{Size} & \textbf{Rate} \\
\midrule
beit-base & microsoft/beit-base-patch16-224 & \cite{bao2021beit} & 350 & 32 & 3.80E-06 \\
beit-large & microsoft/beit-large-patch16-224 & \cite{bao2021beit} & 1259 & 64 & 3.72E-05 \\
deit-tiny & facebook/deit-tiny-patch16-224 & \cite{touvron2021training} & 23 & 16 & 3.63E-05 \\
deit-base & facebook/deit-base-patch16-224 & \cite{touvron2021training} & 346 & 16 & 4.14E-05 \\
dinov2-small & facebook/dinov2-small & \cite{oquab2023dinov2} & 88 & 16 & 5.61E-06 \\
focalnet-tiny & microsoft/focalnet-tiny & \cite{yang2022focal} & 114 & 16 & 5.61E-06 \\
focalnet-base & microsoft/focalnet-base & \cite{yang2022focal} & 353 & 32 & 8.41E-05 \\
resnet-18 & microsoft/resnet-18 & \cite{he2016} & 46 & 16 & 3.63E-05 \\
resnet-50 & microsoft/resnet-50 & \cite{he2016} & 103 & 16 & 9.26E-05 \\
resnet-101 & microsoft/resnet-101 & \cite{he2016} & 167 & 16 & 9.26E-05 \\
vit-base & google/vit-base-patch16-224-in21k & \cite{wu2020visual} & 1198 & 32 & 1.15E-05 \\
vit-large & google/vit-large-patch16-224-in21k & \cite{wu2020visual} & 1249 & 64 & 2.78E-05 \\
\bottomrule
\end{tabularx}
\end{table*}

\footnotetext[1]{Reference for Hugging Face Model Hub  \cite{website:HuggingFace}.}
\footnotetext[2]{Size of PyTorch model binary file.}

\section{Computational Methods}
\label{sec:computational-methods}
\paragraph{} As outlined in Section \ref{sec:tape-laying-process}, the height profiles of tape layers, as captured in the process, are interpretable as image data. Our goal is to develop an AI model with supervised machine learning that has the task of classifying these images into the error categories "gap", "overlap" and "nominal", as previously described in section \ref{sec:tape-laying-process}. This is a classic challenge for image classification in the field of computer vision. A variety of pre-trained models are available for this purpose.

\subsection{Pre-Trained Models}
Deep Learning in Computer Vision remains a dynamic research area, with state-of-the-art models continuously evolving. For a specific use case, it is challenging to predict the most promising architecture \cite{Zhou2021}. In order to reflect the current diversity, we will choose our pre-traines models from the following architectures:

\paragraph{Convolutional Neural Networks (CNNs)} The 2010s were a period of groundbreaking advances in computer vision, initially reflected in the ImageNet Large Scale Visual Recognition Challenge (ILSVRC) leaderboards for object recognition and image classification \cite{ILSVRC15}. CNNs initially proved to be a superior approach, like VGG \cite{simonyan2015a}, ResNet \cite{he2016} and AlexNet \cite{Krizhevsky2012}. Inspired by the biological function of the visual cortex \cite{hubel1962}, the fundamental architecture was established in the 1980s and has been continuously developed \cite{fukushima1980, lecun1989, lecun1998}. Due to their computationally intensive nature, a breakthrough was achieved only with the adaptation of the algorithms to the powerful GPUs available in the 2010s \cite{ciresan2010, ciresan2011}.

\paragraph{Vision Transformers (ViT)} ViTs represent another promising approach in deep learning. Initially designed for text processing \cite{vaswani2017attention}, the core concept of the attention mechanism was successfully adapted for image analysis. These resulting ViTs are adept at recognizing global structures in images and have swiftly outperformed state-of-the-art ResNet-based baselines in challenging datasets \cite{dosovitskiy2020image}.

\paragraph{Focal Modulation Networks (FMNs)} Recent developments have introduced a novel approach, replacing the transformer's self-attention mechanism with a focal modulation module that leverages convolutional layers for modeling token interactions in vision \cite{yang2022focal}. This FMN architecture ranks among the latest state-of-the-art models, demonstrating superior performance over modern attention-based models like BEIT-3 \cite{beit3} on the COCO dataset \cite{lin2015microsoft}.

\paragraph{Open Source} As motivated in section \ref{sec:introduction}, we exclusively utilize pre-trained open-source models. In particular, we employed models from the Hugging Face Model Hub \cite{website:HuggingFace}. The selected model names, architectures, and checkpoints are detailed in Table \ref{tab:models}.

\subsection{Data Preprocessing}
\paragraph{Training, Evaluation and Test Data} Our dataset is randomly partitioned into a training set and a test set. We allocate 70\% of the data for training purposes, while reserving 30\% to uniformly assess the performance of the resulting model. Furthermore, from the training data, an additional 10\% is set aside for evaluation purposes during the training process. This approach ensures that the random split maintains the distribution of categories as described in section  \ref{sec:tape-laying-process}. 

\paragraph{Balancing} As described in section \ref{sec:tape-laying-process}, the dataset has an unbalanced distribution. To prevent the model from favoring the dominant category, the training set is balanced. We simply select an equal number of examples from each class, i.e. "nominal", "gap" and "overlap", to ensure a more even representation across all categories. Thus, we improve the classifier's ability to make accurate predictions across all categories without overfitting to the most prevalent category.

\paragraph{Image Preprocessing} The pre-trained models described in Table \ref{tab:models} were pre-trained on color image datasets, such as ImageNet \cite{ILSVRC15}. Consequently, they typically expect three two-dimensional 8-bit arrays of a specific size, commonly 224x224 pixels. These three channels correspond to the red, green, and blue color channels, with each 8-bit channel capable of representing values ranging from 0 to 255. However, the images from the dataset described in Section \ref{sec:tape-laying-process} are 16-bit grayscale images of size 150x100 pixels. These images have only one channel with values ranging from 0 to 65,536. To apply the models to our dataset, the images need to be appropriately pre-processed. First, the grayscale TIFF images are converted into two-dimensional 8-bit arrays. During this process, no normalization is performed, which would otherwise distort the physical height profile of the images. Instead, the average value of individual pixels is subtracted from all pixel values, and values below 0 or above 255 are appropriately capped. In a second step, this array is duplicated, and the required three color channels are created. Finally, the images must be resized to match the input size of the models. The models we selected expect images of 224x224 pixels or larger. Again, we avoid scaling to prevent distortion of the physical information in the images. Instead, the images are padded with zeros at the edges to reach the required size. We can particularly forego normalization in this case, as the height difference between the substrate and the top surface of the tape laying, as described in Figure \ref{fig:example-tiff}, can be represented within the 255-pixel range.

\paragraph{Data Augmentation} Despite the limited size of our dataset, particularly in the smallest category examples, we have chosen to forego additional data augmentation methods. This decision is based on two considerations: Firstly, we aim to ensure the transferability of our results to other application cases where data augmentation may not be an option. Secondly, there are studies indicating that aggressive data augmentation techniques, such as noising, flipping, and rotation, can significantly impact the performance of pre-trained models \cite {Zoph2020}.

\subsection{Hyperparameter Search}
\paragraph{} For each selected model, a hyperparameter search is conducted to optimize the choice of learning rate and batch size during training. To reduce computational effort, two measures are implemented. Firstly, this is conducted only once beforehand for each model. For this purpose, the training dataset introduced is reduced to 512 examples per category, while the evaluation dataset is retained. Secondly, we employ the Ray Tune \cite{liaw2018} implementation of the Asynchronous Successive Halving Algorithm (ASHA) \cite{li2020} for hyperparameter tuning. The search space for the hyperparameter search is described in Table \ref{tab:hyperparameter-search}, and the settings for ASHA are in Table \ref{tab:asha}. The results of the hyperparameter search are presented in Table \ref{tab:models}.

\begin{table}[h]
\centering
\caption{Parameters for Hyperparameter Search Space.}
\label{tab:hyperparameter-search}
\begin{tabular}{p{0.55\columnwidth} p{0.3\columnwidth}}
\hline
\textbf{Parameter} & \textbf{Values} \\ \hline
Learning Rates & [1e-6, 1e-4] \\
Batch Sizes & [16, 32, 64, 128\footnotemark] \\ \hline
\end{tabular}
\end{table}

\footnotetext{The search was conducted on three Nvidia A6000 GPUs with two trials per GPU. For larger models, this batch size was removed due to memory constraints.}

\begin{table}[h]
\centering
\caption{Parameters for Ray Tune's  \cite{liaw2018} implementation of Asynchronous Successive Halving Algorithm (ASHA) \cite{li2020}.}
\label{tab:asha}
\begin{tabular}{p{0.55\columnwidth} p{0.3\columnwidth}}
\hline
\textbf{Parameter} & \textbf{Values} \\ \hline
Max T & 32 \\
Grace Period & 4 \\
Reduction Factor & 2 \\
Number of Trials & 64 \\ \hline
\end{tabular}
\end{table}

\subsection{Fine-Tuning and Evaluation}
\paragraph{} Using the hyperparameters listed in Table \ref{tab:models}, we train the selected models on a balanced dataset that includes 200, 400, 600, 800, 1200, 1600 and 2000 instances for each category "gap", "nominal" and "overlap". Following the hyperparameter optimization procedure, the evaluation dataset is kept without modification. After fine-tuning, we evaluate the performance of the models using the previously described test dataset. Since our problem is a multi-class problem, we calculate the mean $\text{F}_1$ score to evaluate the classification efficiency. The outcomes of these calculations are illustrated in Figure \ref{fig:f1-score-vs-examples}.

\section{Results}
\label{sec:results}
\paragraph{} We aim to evaluate the performance of the fine-tuned models from Table \ref{tab:models} across varying sizes of training datasets, that is, under a defined number of available examples per category 'gap', 'nominal', and 'overlap'.

\subsection{Impact of Architecture}
\begin{figure}[ht]
\begin{center}
\centerline{\includegraphics[width=\columnwidth]{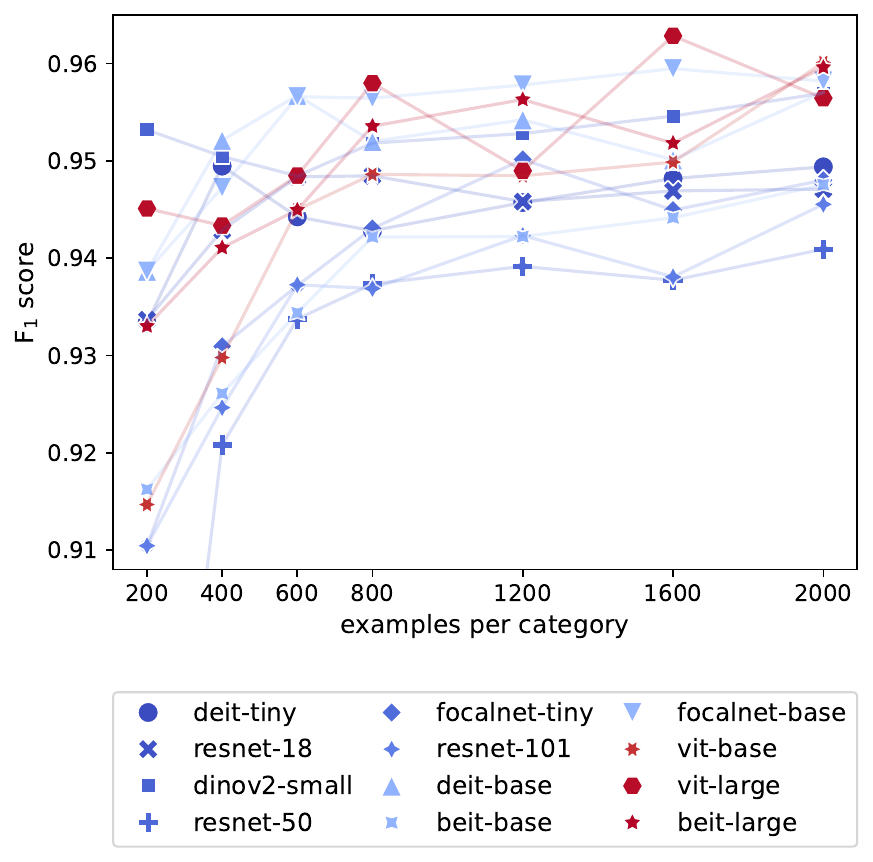}}
\caption{Analysis of sensitivity to training dataset size for fine-tuned models from table \ref{tab:models}: Averaged $\text{F}_1$ score versus number of examples per category in the training dataset. The results are colored in relation to the PyTorch model binary file sizes in a continuous color gradient from blue (small) to red (large).}
\label{fig:f1-score-vs-examples}
\end{center}
\end{figure}

\paragraph{} To evaluate the impact of the architecture on the performance of the fine-tuned models, in Figure \ref{fig:f1-score-vs-examples} we investigate the $\text{F}_1$ score of each model in relation to the provided examples per category.

\paragraph{Area of Stable Performance} We observe that under the largest dataset with 2000 examples, all models show a good $\text{F}_1$ score. The weakest model, the CNN ResNet-18, scores approximately 0.94 points, only about 0.02 points below the best model, vit-base, which scores around 0.96 points. Upon closer differentiation, a leading group emerges, showing only a 0.005 point difference among themselves. This group exclusively includes ViTs (dinov2-small, vit-base, vit-Large, deit-base, and beit-large) and one FMN (focalnet-base). A chasing group of models with ViTs (deit-tiny, beit-base), FMNs (focalnet-tiny), and CNNs (resnet-18, resnet-50, resnet-101) is clustered about 0.01 point below. With a continuous reduction in the number of examples per category available for fine-tuning, these two groups form relatively well-separated bands with stable performance down to a threshold of 800 examples.

\paragraph{Area of Performance Decline} When the threshold of 800 examples is crossed, nearly all models experience an accelerating decline in the $\text{F}_1$ score. For the leading group, this decline is less pronounced at approximately 0.02 points, compared to the following group, which sees a decline of approximately 0.03 points. The groups largely remain intact, with only the two ViTs deit-tiny and vit-base and the CNN resnet-18 changing bands. The ViT dinov2-Small proves to be particularly stable, showing no significant loss in comparison to having 200 examples.

\subsection{Influence of model size}
\begin{figure*}[ht]
\begin{center}
    \begin{tabular}{cc}
        \includegraphics[width=0.45\textwidth]{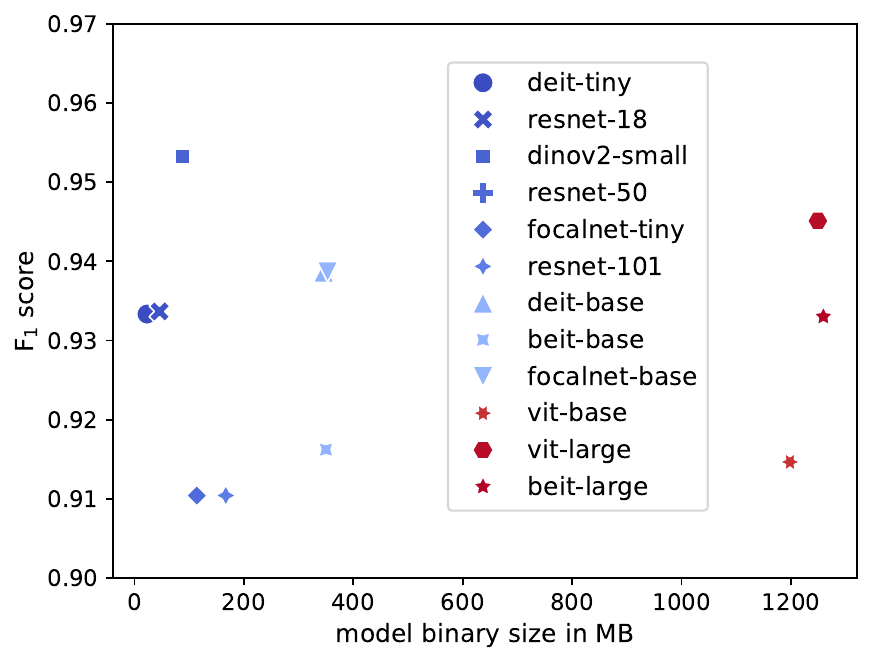} &
        \includegraphics[width=0.45\textwidth]{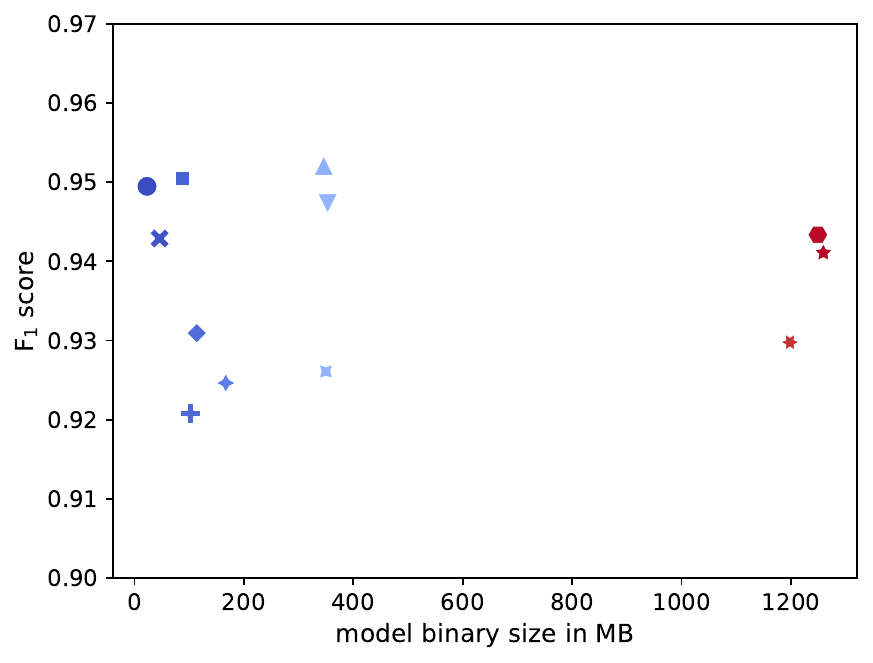} \\
        (a) 200 examples per category. &
        (b) 400 examples per category. \\
        \includegraphics[width=0.45\textwidth]{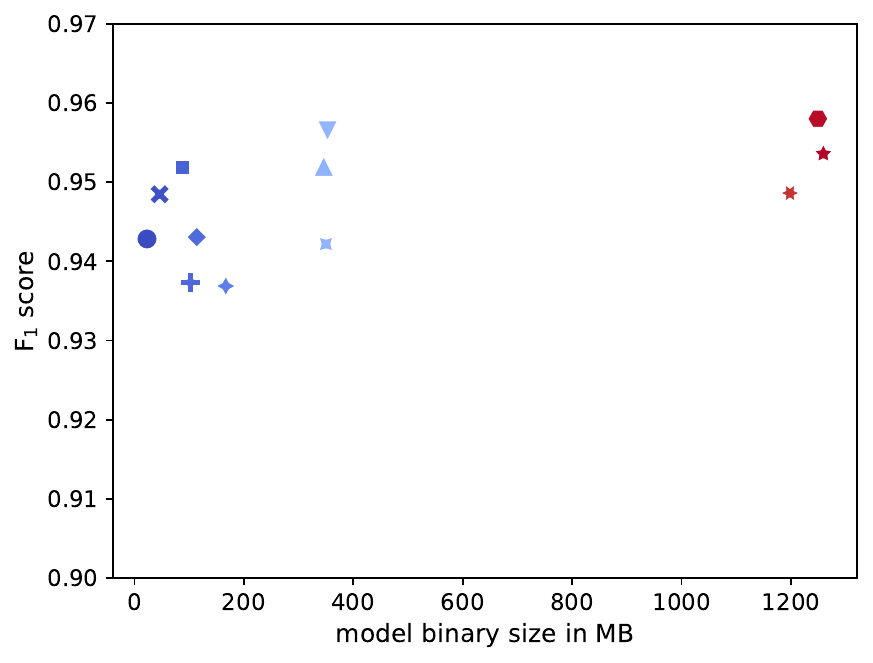} &
        \includegraphics[width=0.45\textwidth]{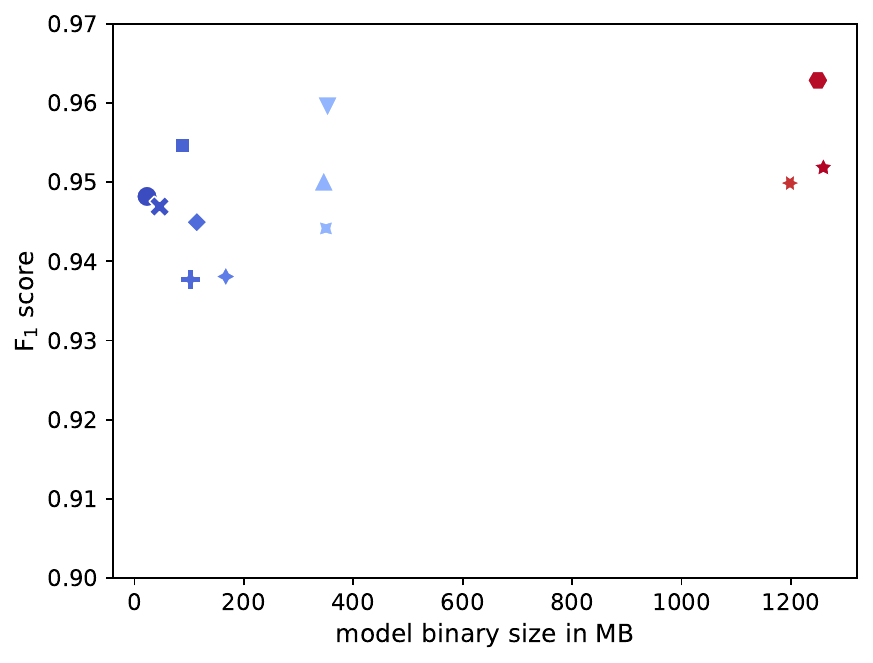} \\
        (c) 800 examples per category. &
        (d) 1600 examples per category. \\
    \end{tabular}
\caption{Analysis of performance relative to model size: Averaged $\text{F}_1$ scores versus PyTorch model binary file sizes for fine-tuned models from Table \ref{tab:models} across four training datasets with diverse category sample sizes.}
\label{fig:f1-score-vs-size}
\end{center}
\end{figure*}

\paragraph{} The previously mentioned transition of resnet-18, the smallest CNN in our selection, from the chasing to the leading group. We aim to examine the different behaviors of larger and smaller models regarding the reduction of data available for fine-tuning in more detail. For this purpose, we utilize the $\text{F}_1$ scores shown in Figure \ref{fig:f1-score-vs-examples} and correlate them with their model sizes noted in Table \ref{tab:models}. We illustrate this in Figure \ref{fig:f1-score-vs-size} by considering two examples above and two examples below the mentioned threshold of 800 examples per category.

\paragraph{Model Groups} We see three groups of models. The previously mentioned group of large models (vit-base, vit-large, and beit-large), a group of medium-sized models (deit-base, beit-base, and focalnet-base), and finally the group of smaller models, among which we particularly want to monitor the subgroup of the smallest models (deit-tiny, resnet-18, and dinov2-small).

\paragraph{Larger Models Strengths} When examining the two representations in Figure \ref{fig:f1-score-vs-examples} with over 800 examples per category, within the mentioned area of stable performance, two key observations emerge. Firstly, all model groups form performance clusters that are closely spaced, which means that their $\text{F}_1$ scores differ by less than 0.02 points. Secondly, there exists a noticeable trend: the larger the model, the better its performance. This observation aligns with the earlier conclusion that the group of larger models tends to dominate within the realm of stable performance.

\paragraph{Small Model Stability} However, when we drop below the mentioned threshold of 800 samples per category, three effects become evident. First, the performance clusters of the three model groups expand, i.e. the variance of the $\text{F}_1$ scores within the groups significantly increases. Second, the trend that larger models yield better results dissipates. The most remarkable, however, is the third effect. Within the group of smaller models, which shows the most increase in variance, the performance of the subgroup of the smallest models remains surprisingly stable. This includes not only the two ViTs, deit-tiny and dinov2-small, but also resnet-18, the smallest CNN in our selection. The $F_1$ score of the ViT model dinov2-small with 200 examples per category is less than 0.01 points below that of the best large model, vit-large, with 1600 examples per category.

\subsection{Practical Application Insights}
\paragraph{Accuracy Evaluation} As outlined in Section \ref{sec:introduction}, the original motivation of our work is to provide an automated quality control system using AI, even with a small number of examples. In practical application, particularly, the accuracy of the prediction of the fine-tuned models is a critical criterion for deployment. We have calculated this based on the test dataset introduced in Section \ref{sec:computational-methods} for the respective categories in Table \ref{tab:acccuracies}. Motivated by previous results, we concentrated on the subgroup of the three smallest models as detailed in Table \ref{tab:models}. These models were fine-tuned using example data from 200 examples per category.

\paragraph{Comparison} To contextualize the quality of these fine-tuned models in relation to existing solutions, results from a comparable study were considered, which were computed with Deep Convolutional Generative Adversarial Network (DCGAN) and Geometrical Transformation (GT) techniques for the CRFP use-case on a different dataset using synthetic image data augmentation \cite{Meister2021}. The results are not directly comparable since they were obtained on a different dataset. However, this allows for a rough assessment of the models calculated here. The nominal and gap categories are predicted with higher accuracy by all models presented here. For overlap, at least resnet-18 surpasses previous accuracies.

\begin{table}[h]
\centering
\caption{Predicted accuracies on test dataset for categories nominal, overlap and gap for the group of smallest models in table \ref{tab:models}.}
\label{tab:acccuracies}
\begin{tabular}{p{0.35\columnwidth}  p{0.15\columnwidth}  p{0.15\columnwidth} p{0.15\columnwidth}}
\hline
\textbf{Short Name} & \textbf{Nominal} & \textbf{Overlap} & \textbf{Gap} \\ \hline
dinov2-small & 95.72\% & 81.15\% & 96.78\% \\
deit-tiny & 92.64\% & 85.76\% & 97.80\% \\
resnet-18 & 92.36\% & 90.27\% & 96.62\% \\ \hline
DCGAN+GT\footnotemark & 88.37\% & 87.23\% & 93.41\% \\
\hline
\end{tabular}
\end{table}

\footnotetext{Accuracies computed with Deep Convolutional Generative Adversarial Network and Geometrical Transformation techniques for CRFP use-case on a different dataset using synthetic image data augmentation \cite{Meister2021}.}

\section{Conclusion}
\label{sec:conclusion}

\paragraph{} Our investigation supports a comprehensive understanding of performance dynamics across different model architectures and sizes, which highlights the applicability of small models to small datasets and establishes the central role of transfer learning for our use case. Here we summarize our key findings:

\paragraph{Model Architectures} Our study shows that ViTs and FMNs generally perform better than Convolutional Neural Networks (CNNs) on a range of datasets sizes. This supports previous research and introduces FMNs as a strong option. Since predicting the best model for a given task is difficult, incorporating a variety of models, such as ViTs and FMNs, in the search for the best hyperparameters is a practical approach.

\paragraph{Model Sizes} Contrary to the common assumption that larger models generally perform better, our research shows that small models exhibit remarkable efficiency, especially when applied to smaller datasets. If this finding is transferable to other applications, it is recommended to experiment with a number of smaller models rather than a few large models when optimizing hyperparameters. This strategy offers two advantages: Saving computational resources and accelerating insights. In addition, the near-equivalent performance of small models, even with only 200 examples, compared to larger models trained with 1600 examples, offers significant advantages for edge computing devices. By enabling local high-quality predictions, small models reduce bandwidth requirements and ensure low latency in fast production environments. In addition, the ability to specialize and combine multiple small models for different fault classifications opens up new ways to achieve exceptional accuracy with minimal resources.

\paragraph{Use-Case Application} The results presented in section \ref{sec:results} confirm the competitiveness, perhaps even the superiority, of transfer learning as a method in contexts constrained by limited data. The fine-tuned models not only demonstrate the feasibility of productive use, but also offer a compelling alternative to manually adapting algorithms. This research  shows that with transfer learning, we are able to develop efficient AI models tailored to specific application requirements even with limited data.

\paragraph{} Our research supports a paradigm shift towards resource-efficient, adaptable and accessible AI model development. By leveraging the strengths of Open Source ViTs, FMNs and the strategic use of transfer learning, we are demonstrating the way for innovative solutions that are not only computationally efficient, but also offer robust performance in a variety of use cases.

\section*{Impact Statement}
\paragraph{} This paper presents work aimed at advancing the field of Machine Learning and CFRP Tape Laying. There are many potential technical and economic consequences of our work in industrial production, but none that we feel must be specifically highlighted here.

\bibliography{paper}

\begin{thebibliography}{28}
\providecommand{\natexlab}[1]{#1}
\providecommand{\url}[1]{\texttt{#1}}
\expandafter\ifx\csname urlstyle\endcsname\relax
  \providecommand{\doi}[1]{doi: #1}\else
  \providecommand{\doi}{doi: \begingroup \urlstyle{rm}\Url}\fi

\bibitem[Bao et~al.(2021)Bao, Dong, and Wei]{bao2021beit}
Bao, H., Dong, L., and Wei, F.
\newblock Beit: {BERT} pre-training of image transformers.
\newblock \emph{CoRR}, abs/2106.08254, 2021.
\newblock URL \url{https://arxiv.org/abs/2106.08254}.

\bibitem[Cireşan et~al.(2011)Cireşan, Meier, Masci, Gambardella, and Schmidhuber]{ciresan2011}
Cireşan, D., Meier, U., Masci, J., Gambardella, L.~M., and Schmidhuber, J.
\newblock High-performance neural networks for visual object classification.
\newblock \emph{Computing Research Repository - CORR}, 02 2011.

\bibitem[Cireşan et~al.(2010)Cireşan, Meier, Gambardella, and Schmidhuber]{ciresan2010}
Cireşan, D.~C., Meier, U., Gambardella, L.~M., and Schmidhuber, J.
\newblock {Deep, Big, Simple Neural Nets for Handwritten Digit Recognition}.
\newblock \emph{Neural Computation}, 22\penalty0 (12):\penalty0 3207--3220, 12 2010.
\newblock ISSN 0899-7667.
\newblock \doi{10.1162/NECO_a_00052}.
\newblock URL \url{https://doi.org/10.1162/NECO\_a\_00052}.

\bibitem[Djolonga et~al.(2020)Djolonga, Yung, Tschannen, Romijnders, Beyer, Kolesnikov, Puigcerver, Minderer, D'Amour, Moldovan, Gelly, Houlsby, Zhai, and Lucic]{Djolonga2020}
Djolonga, J., Yung, J., Tschannen, M., Romijnders, R., Beyer, L., Kolesnikov, A., Puigcerver, J., Minderer, M., D'Amour, A., Moldovan, D., Gelly, S., Houlsby, N., Zhai, X., and Lucic, M.
\newblock On robustness and transferability of convolutional neural networks.
\newblock \emph{CoRR}, abs/2007.08558, 2020.
\newblock URL \url{https://arxiv.org/abs/2007.08558}.

\bibitem[Dosovitskiy et~al.(2021)Dosovitskiy, Beyer, Kolesnikov, Weissenborn, Zhai, Unterthiner, Dehghani, Minderer, Heigold, Gelly, Uszkoreit, and Houlsby]{dosovitskiy2020image}
Dosovitskiy, A., Beyer, L., Kolesnikov, A., Weissenborn, D., Zhai, X., Unterthiner, T., Dehghani, M., Minderer, M., Heigold, G., Gelly, S., Uszkoreit, J., and Houlsby, N.
\newblock An image is worth 16x16 words: Transformers for image recognition at scale.
\newblock In \emph{9th International Conference on Learning Representations, {ICLR} 2021, Virtual Event, Austria, May 3-7, 2021}. OpenReview.net, 2021.
\newblock URL \url{https://openreview.net/forum?id=YicbFdNTTy}.

\bibitem[Fukushima(1980)]{fukushima1980}
Fukushima, K.
\newblock Neocognitron: A self-organizing neural network model for a mechanism of pattern recognition unaffected by shift in position.
\newblock \emph{Biological Cybernetics}, 36\penalty0 (4):\penalty0 193--202, Apr 1980.
\newblock ISSN 1432-0770.
\newblock \doi{10.1007/BF00344251}.
\newblock URL \url{https://doi.org/10.1007/BF00344251}.

\bibitem[He et~al.(2016)He, Zhang, Ren, and Sun]{he2016}
He, K., Zhang, X., Ren, S., and Sun, J.
\newblock Deep residual learning for image recognition.
\newblock In \emph{2016 IEEE Conference on Computer Vision and Pattern Recognition (CVPR)}, pp.\  770--778, 2016.
\newblock \doi{10.1109/CVPR.2016.90}.

\bibitem[Hubel \& Wiesel(1962)Hubel and Wiesel]{hubel1962}
Hubel, D.~H. and Wiesel, T.~N.
\newblock Receptive fields, binocular interaction and functional architecture in the cat's visual cortex.
\newblock \emph{The Journal of Physiology}, 160\penalty0 (1):\penalty0 106--154, 1962.
\newblock \doi{https://doi.org/10.1113/jphysiol.1962.sp006837}.
\newblock URL \url{https://physoc.onlinelibrary.wiley.com/doi/abs/10.1113/jphysiol.1962.sp006837}.

\bibitem[{Hugging Face}(2024)]{website:HuggingFace}
{Hugging Face}.
\newblock {Model Hub}, 2024.
\newblock URL \url{{https://huggingface.co/models}}.

\bibitem[Kolesnikov et~al.(2019)Kolesnikov, Beyer, Zhai, Puigcerver, Yung, Gelly, and Houlsby]{Kolesnikov2019}
Kolesnikov, A., Beyer, L., Zhai, X., Puigcerver, J., Yung, J., Gelly, S., and Houlsby, N.
\newblock Large scale learning of general visual representations for transfer.
\newblock \emph{CoRR}, abs/1912.11370, 2019.
\newblock URL \url{http://arxiv.org/abs/1912.11370}.

\bibitem[Krizhevsky et~al.(2012)Krizhevsky, Sutskever, and Hinton]{Krizhevsky2012}
Krizhevsky, A., Sutskever, I., and Hinton, G.~E.
\newblock Imagenet classification with deep convolutional neural networks.
\newblock In Pereira, F., Burges, C., Bottou, L., and Weinberger, K. (eds.), \emph{Advances in Neural Information Processing Systems}, volume~25. Curran Associates, Inc., 2012.

\bibitem[LeCun et~al.(1989)LeCun, Boser, Denker, Henderson, Howard, Hubbard, and Jackel]{lecun1989}
LeCun, Y., Boser, B., Denker, J.~S., Henderson, D., Howard, R.~E., Hubbard, W., and Jackel, L.~D.
\newblock Backpropagation applied to handwritten zip code recognition.
\newblock \emph{Neural Computation}, 1\penalty0 (4):\penalty0 541--551, 1989.
\newblock \doi{10.1162/neco.1989.1.4.541}.

\bibitem[Lecun et~al.(1998)Lecun, Bottou, Bengio, and Haffner]{lecun1998}
Lecun, Y., Bottou, L., Bengio, Y., and Haffner, P.
\newblock Gradient-based learning applied to document recognition.
\newblock \emph{Proceedings of the IEEE}, 86\penalty0 (11):\penalty0 2278--2324, 1998.
\newblock \doi{10.1109/5.726791}.

\bibitem[Li et~al.(2018)Li, Jamieson, Rostamizadeh, Gonina, Hardt, Recht, and Talwalkar]{li2020}
Li, L., Jamieson, K.~G., Rostamizadeh, A., Gonina, E., Hardt, M., Recht, B., and Talwalkar, A.
\newblock Massively parallel hyperparameter tuning.
\newblock \emph{CoRR}, abs/1810.05934, 2018.
\newblock URL \url{http://arxiv.org/abs/1810.05934}.

\bibitem[Liaw et~al.(2018)Liaw, Liang, Nishihara, Moritz, Gonzalez, and Stoica]{liaw2018}
Liaw, R., Liang, E., Nishihara, R., Moritz, P., Gonzalez, J.~E., and Stoica, I.
\newblock Tune: A research platform for distributed model selection and training.
\newblock \emph{ArXiv}, abs/1807.05118, 2018.
\newblock URL \url{https://api.semanticscholar.org/CorpusID:49741140}.

\bibitem[Lin et~al.(2014)Lin, Maire, Belongie, Hays, Perona, Ramanan, Doll{\'a}r, and Zitnick]{lin2015microsoft}
Lin, T.-Y., Maire, M., Belongie, S., Hays, J., Perona, P., Ramanan, D., Doll{\'a}r, P., and Zitnick, C.~L.
\newblock Microsoft coco: Common objects in context.
\newblock In Fleet, D., Pajdla, T., Schiele, B., and Tuytelaars, T. (eds.), \emph{Computer Vision -- ECCV 2014}, pp.\  740--755, Cham, 2014. Springer International Publishing.
\newblock ISBN 978-3-319-10602-1.

\bibitem[Meister et~al.(2021)Meister, M{\"o}ller, St{\"u}ve, and Groves]{Meister2021}
Meister, S., M{\"o}ller, N., St{\"u}ve, J., and Groves, R.~M.
\newblock Synthetic image data augmentation for fibre layup inspection processes: Techniques to enhance the data set.
\newblock \emph{Journal of Intelligent Manufacturing}, 32\penalty0 (6):\penalty0 1767--1789, Aug 2021.
\newblock ISSN 1572-8145.
\newblock \doi{10.1007/s10845-021-01738-7}.
\newblock URL \url{https://doi.org/10.1007/s10845-021-01738-7}.

\bibitem[Oquab et~al.(2023)Oquab, Darcet, Moutakanni, Vo, Szafraniec, Khalidov, Fernandez, Haziza, Massa, El-Nouby, Assran, Ballas, Galuba, Howes, Huang, Li, Misra, Rabbat, Sharma, Synnaeve, Xu, Jegou, Mairal, Labatut, Joulin, and Bojanowski]{oquab2023dinov2}
Oquab, M., Darcet, T., Moutakanni, T., Vo, H., Szafraniec, M., Khalidov, V., Fernandez, P., Haziza, D., Massa, F., El-Nouby, A., Assran, M., Ballas, N., Galuba, W., Howes, R., Huang, P.-Y., Li, S.-W., Misra, I., Rabbat, M., Sharma, V., Synnaeve, G., Xu, H., Jegou, H., Mairal, J., Labatut, P., Joulin, A., and Bojanowski, P.
\newblock Dinov2: Learning robust visual features without supervision, 2023.

\bibitem[Russakovsky et~al.(2014)Russakovsky, Deng, Su, Krause, Satheesh, Ma, Huang, Karpathy, Khosla, Bernstein, Berg, and Fei{-}Fei]{ILSVRC15}
Russakovsky, O., Deng, J., Su, H., Krause, J., Satheesh, S., Ma, S., Huang, Z., Karpathy, A., Khosla, A., Bernstein, M.~S., Berg, A.~C., and Fei{-}Fei, L.
\newblock Imagenet large scale visual recognition challenge.
\newblock \emph{CoRR}, abs/1409.0575, 2014.

\bibitem[Schuster et~al.(2020)Schuster, Mayer, Willmeroth, Brandt, and Kupke]{dlr137804}
Schuster, A., Mayer, M., Willmeroth, M., Brandt, L., and Kupke, M.
\newblock Inline quality control for thermoplastic automated fibre placement.
\newblock \emph{Procedia Manufacturing}, 51:\penalty0 505--511, 2020.
\newblock ISSN 2351-9789.
\newblock \doi{https://doi.org/10.1016/j.promfg.2020.10.071}.
\newblock URL \url{https://www.sciencedirect.com/science/article/pii/S2351978920319284}.
\newblock 30th International Conference on Flexible Automation and Intelligent Manufacturing (FAIM2021).

\bibitem[Simonyan \& Zisserman(2015)Simonyan and Zisserman]{simonyan2015a}
Simonyan, K. and Zisserman, A.
\newblock Very deep convolutional networks for large-scale image recognition.
\newblock In \emph{3rd International Conference on Learning Representations}, pp.\  1--14. Computational and Biological Learning Society, 2015.

\bibitem[Touvron et~al.(2021)Touvron, Cord, Douze, Massa, Sablayrolles, and Jégou]{touvron2021training}
Touvron, H., Cord, M., Douze, M., Massa, F., Sablayrolles, A., and Jégou, H.
\newblock Training data-efficient image transformers \& distillation through attention, 2021.

\bibitem[Vaswani et~al.(2017)Vaswani, Shazeer, Parmar, Uszkoreit, Jones, Gomez, Kaiser, and Polosukhin]{vaswani2017attention}
Vaswani, A., Shazeer, N., Parmar, N., Uszkoreit, J., Jones, L., Gomez, A.~N., Kaiser, L.~u., and Polosukhin, I.
\newblock Attention is all you need.
\newblock In Guyon, I., Luxburg, U.~V., Bengio, S., Wallach, H., Fergus, R., Vishwanathan, S., and Garnett, R. (eds.), \emph{Advances in Neural Information Processing Systems}, volume~30. Curran Associates, Inc., 2017.

\bibitem[Wang et~al.(2023)Wang, Bao, Dong, Bjorck, Peng, Liu, Aggarwal, Mohammed, Singhal, Som, and Wei]{beit3}
Wang, W., Bao, H., Dong, L., Bjorck, J., Peng, Z., Liu, Q., Aggarwal, K., Mohammed, O.~K., Singhal, S., Som, S., and Wei, F.
\newblock Image as a foreign language: Beit pretraining for vision and vision-language tasks.
\newblock In \emph{Proceedings of the IEEE/CVF Conference on Computer Vision and Pattern Recognition (CVPR)}, pp.\  19175--19186, June 2023.

\bibitem[Wu et~al.(2020)Wu, Xu, Dai, Wan, Zhang, Yan, Tomizuka, Gonzalez, Keutzer, and Vajda]{wu2020visual}
Wu, B., Xu, C., Dai, X., Wan, A., Zhang, P., Yan, Z., Tomizuka, M., Gonzalez, J., Keutzer, K., and Vajda, P.
\newblock Visual transformers: Token-based image representation and processing for computer vision, 2020.

\bibitem[Yang et~al.(2022)Yang, Li, Dai, and Gao]{yang2022focal}
Yang, J., Li, C., Dai, X., and Gao, J.
\newblock Focal modulation networks.
\newblock In Oh, A.~H., Agarwal, A., Belgrave, D., and Cho, K. (eds.), \emph{Advances in Neural Information Processing Systems}, 2022.
\newblock URL \url{https://openreview.net/forum?id=ePhEbo039l}.

\bibitem[Zhou et~al.(2021)Zhou, Lu, Yang, and Yu]{Zhou2021}
Zhou, H.-Y., Lu, C., Yang, S., and Yu, Y.
\newblock Convnets vs. transformers: Whose visual representations are more transferable?
\newblock In \emph{2021 IEEE/CVF International Conference on Computer Vision Workshops (ICCVW)}, pp.\  2230--2238, 2021.
\newblock \doi{10.1109/ICCVW54120.2021.00252}.

\bibitem[Zoph et~al.(2020)Zoph, Ghiasi, Lin, Cui, Liu, Cubuk, and Le]{Zoph2020}
Zoph, B., Ghiasi, G., Lin, T.-Y., Cui, Y., Liu, H., Cubuk, E.~D., and Le, Q.~V.
\newblock Rethinking pre-training and self-training.
\newblock NIPS'20, Red Hook, NY, USA, 2020. Curran Associates Inc.
\newblock ISBN 9781713829546.

\end{thebibliography}
\bibliographystyle{icml2024}


\newpage
\appendix
\onecolumn
\section{Code and Data Accessibility}

For the purpose of review and to facilitate reproducibility of our findings, we have provided an anonymized version of our code repository. It can be accessed through the following link:

\textit{https://anonymous.4open.science/r/crfp-transfer-learning/README.md}

We encourage reviewers and interested parties to consult the README file within the repository for detailed instructions on how to access and utilize the code. For accessing the necessary dataset, please use the provided S3 credentials:

\begin{verbatim}
AWS_ACCESS_KEY_ID="GOOG1E7FETDFKZD3RQB75RPGQAXOELXNY4QEKLFO3J3BM2NTVGO4HSROJ24IA"
AWS_SECRET_ACCESS_KEY="PhQK7ymyI6UDaxeytOyWr+AiHCQOnDKgFl/5KMTH"
\end{verbatim}

It is our intention to publish the code repository and make the selected models and data publicly available on GitHub and Hugging Face, respectively, upon conclusion of the review process. This will ensure that our contributions to the field are accessible for further research and application in the broader AI and machine learning community.

\end{document}